\documentclass[12pt]{iopart}

\usepackage{iopams}  
\usepackage{graphicx}
\usepackage{algorithm}
\usepackage{algpseudocode}
\begin{document}

\title[Molecular Reinforcement Learning with Adaptive Intrinsic Rewards]{Mol-AIR: Molecular Reinforcement Learning with Adaptive Intrinsic Rewards for Goal-directed Molecular Generation}

\author{Jinyeong Park$^1$, Jaegyoon Ahn$^1$, Jonghwan Choi$^{2,*}$, and Jibum Kim$^{1,3,*}$}

\address{$^1$ Department of Computer Science and Engineering, Incheon National University, Yeonsu-gu, Incheon, Republic of Korea}
\address{$^2$ College of Information Science, Hallym University, Chuncheon, Gangwon-do, Republic of Korea}
\address{$^3$ Center for Brain-Machine Interface, Incheon National University, Yeonsu-gu, Incheon, Republic of Korea}
\address{$^*$ These authors are co-corresponding authors.}

\eads{\mailto{jonghwanc@hallym.ac.kr}, \mailto{jibumkim@inu.ac.kr}}

\vspace{10pt}
\begin{indented}
\item[]March 2024
\end{indented}

\begin{abstract}
Optimizing techniques for discovering molecular structures with desired properties is crucial in artificial intelligence(AI)-based drug discovery. Combining deep generative models with reinforcement learning has emerged as an effective strategy for generating molecules with specific properties. Despite its potential, this approach is ineffective in exploring the vast chemical space and optimizing particular chemical properties. To overcome these limitations, we present Mol-AIR, a reinforcement learning-based framework using adaptive intrinsic rewards for effective goal-directed molecular generation. Mol-AIR leverages the strengths of both history-based and learning-based intrinsic rewards by exploiting random distillation network and counting-based strategies. In benchmark tests, Mol-AIR demonstrates superior performance over existing approaches in generating molecules with desired properties without any prior knowledge, including penalized LogP, QED, and celecoxib similarity. We believe that Mol-AIR represents a significant advancement in drug discovery, offering a more efficient path to discovering novel therapeutics.
\end{abstract}

%
\vspace{2pc}
\maketitle
%
%

\section{Introduction}
\label{sec:introduction}

The development of optimization techniques to efficiently discover molecular structures with target properties is a critical challenge in artificial intelligence (AI)-based drug discovery research. In traditional drug discovery, high-throughput screening (HTS) techniques, which investigate the properties of compounds in large chemical libraries, are used to identify hit molecules that exhibit desired pharmacological properties\cite{pereira2007origin}. However, the HTS approach has limitations in reducing the time and costs of hit discovery\cite{kumar2022machine}. Additionally, the number of drug-like compounds is estimated to be in the range of ${10}^{33}$ to ${10}^{60}$\cite{polishchuk2013estimation}, making it difficult and time-consuming to discover desired hit and lead molecules against target diseases.

Deep generative models have significantly advanced, and these advancements have been applied to the efficient and effective exploration of molecular structures in drug discovery\cite{chen2023artificial}. While HTS involves selecting hits from a known vast chemical library, an approach using deep generative models generates hits directly by creating molecular structures with target properties, a process called goal-directed molecular generation\cite{brown2019guacamol, bilodeau2022generative}. Goal-directed molecular generation faces two key challenges: 1) representing and generating molecular structures using a deep generative model and 2) directing a deep generative model to discover molecules possessing desired chemical properties. To address these challenges, many studies have used the simplified molecular-input line-entry system (SMILES), self-referencing embedded strings (SELFIES)\cite{krenn2022selfies, jones2023molecular}, and graph-based representation methods\cite{jin2018junction} in training deep molecular generative models. Researchers have exploited various deep generative models, including recurrent neural networks (RNNs)\cite{choi2023rebadd, kim2022predicting}, transformers\cite{bagal2021molgpt}, and graph neural networks (GNNs)\cite{jin2018junction}, to efficiently handle those string-based or graph-based molecular data\cite{cheng2021molecular, deng2022artificial}. Furthermore, Bayesian optimization\cite{jin2018junction, gomez2018automatic} and reinforcement learning (RL) techniques\cite{choi2023rebadd, olivecrona2017molecular, zhavoronkov2019deep} have been exploited for deep molecular generative models to create molecules with desired chemical properties.

Many studies have demonstrated the effectiveness of molecular structure generation strategies using RL for optimizing various molecular structure properties\cite{choi2023rebadd, kim2022predicting, olivecrona2017molecular, zhavoronkov2019deep, pereira2021diversity}. The typical RL configuration in AI-based drug design consists of two components: an agent that generates molecular structures and an environment that evaluate the generated molecules. If a generated molecular structure possesses desired target properties, such as a high quantitative estimate of drug-likeness (QED), the agent receives a high reward as feedback from the environment. These property-constrained RL methods are effective in fine-tuning a pre-trained molecular generative model so that it can generate a large number of hit molecules\cite{choi2023rebadd, kim2022predicting, olivecrona2017molecular, zhavoronkov2019deep, pereira2021diversity}. However, owing to the vast size of the chemical space, it is challenging for an agent to perform efficient exploration, resulting in failure to find an optimal policy for the desired generation of molecular structures\cite{deng2022artificial, yang2021hit, thiede2022curiosity}.

To enhance the exploration capability of property-constrained RL, curiosity strategies have been recently proposed\cite{thiede2022curiosity, chadi2023curiosity}. In the curiosity-based RL scheme, there are two types of rewards: extrinsic and intrinsic rewards. An extrinsic reward, aimed at improving target chemical properties, is calculated based on evaluated property scores, whereas an intrinsic reward is used to enhance exploration ability and find new molecular structures without relying on target properties. Intrinsic rewards encourage the learning of diverse molecular structures and the discovery of hits with higher target properties. In \cite{thiede2022curiosity}, the authors demonstrated the effectiveness of using intrinsic rewards to optimize chemical properties, such as penalized LogP (pLogP) and QED. However, existing intrinsic reward methods are not effective in tasks involving the generation of compounds structurally similar to specific drugs (e.g., celecoxib). Furthermore, as intrinsic reward is calculated by predefined algorithms, it is necessary to heuristically adjust those calculation algorithms to apply it for various chemical property optimizations.

In this study, we propose a new RL-based framework using a novel intrinsic reward method that can improve the exploration ability of RL for various chemical properties. To the best of our knowledge, the proposed framework is the first molecular optimization framework utilizing a combination of two types of intrinsic rewards based on a random distillation network (RND)\cite{burda2018exploration} and counting-based strategies. Compared to existing intrinsic reward functions, our method demonstrated superior performance in goal-directed molecular generation without any prior knowledge for six chemical properties, including pLogP, QED, and drug similarity. The proposed framework significantly outperformed the existing approach in tasks related to identifying hit molecules with structural similarities. Furthermore, we investigated and explained the superiority of the proposed framework by performing an ablation study and hyperparameter analysis.

\section{Preliminaries}
\label{sec:preliminaries}

\subsection{Molecular Structure Representation}
\label{subsec:molecular_structure_representation}
The selection of molecular structure representation methods is crucial in the field of AI-driven drug discovery\cite{david2020molecular}. The fundamental elements of molecular structures are atoms and the bonds between them, which are clearly described by both string-based and graph-based representation methods. In graph-based representations, atoms are denoted as nodes and their connections as edges. This method excellently captures the overall structure of compounds, yet it faces the challenge of lacking a standardized convention for configuring node and edge features. Moreover, the choice of a molecular graph representation often hinges on the specific graph traversal algorithm used, rendering the selection of representation a task-specific consideration\cite{david2020molecular}.

SMILES and SELFIES are widely recognized as standard molecular representation methods in string-based molecular generation tasks\cite{krenn2022selfies, jones2023molecular}. Both SMILES and SELFIES represent atomic information through alphabet characters and bond information using symbols such as "-", "=", and "\#". The key difference between these two string-representation methods is in their handling of branching and ring structures. While SMILES uses parentheses to denote branches and numeric notations for ring closures, SELFIES utilizes specialized symbols like [Branch] and [Ring] for these purposes.

Compared to SMILES, which requires accurate closure of parentheses and precise ring numbering, SELFIES provides several advantages in the training of molecular generative models. It enables easier management of errors that may occur in generated molecular structure strings through predefined rule-based algorithms\cite{krenn2022selfies}. Its inherent flexibility and error-handling capability allow more effective navigation of structural variations and correction of errors, thereby enhancing the efficiency of model training\cite{krenn2022selfies, cheng2023group}. Consequently, we used SELFIES to address the challenges of generating valid molecular structures and to enhance reinforcement learning's ability to efficiently explore the vast chemical space.

\subsection{Reinforcement Learning}
\label{subsec:reinforcement_learning}
In RL for goal-directed molecular generation, a Markov decision process (MDP) $\mathcal{M}=(\mathcal{S},\mathcal{A},p,r,\gamma)$ is defined as follows: $\mathcal{S}$ is a set of states representing SELFIES strings, $\mathcal{A}$ is a set of actions representing SELFIES characters, $p(s_{t+1}|s_t,a_t)$ is a state transition probability distribution for a next SELFIES character, $r(s_t,a_t)$ denotes the reward function which provides a scalar reward $r_t$ when action $a_t$ is executed in state $s_t$, and $\gamma\in\left(0,1\right]$ is a scalar discount factor. The goal of RL is to discover a policy $\pi\left(a_t|s_t\right)$ that maximizes the expected sum of discounted rewards, formalized as the objective function $J(\pi)=\mathbb{E}\left[\sum_{k=t}^{T}{\gamma^{k-t}r_k}\right]$.

Proximal Policy Optimization (PPO)\cite{schulman2017proximal}, a notable policy gradient algorithm, enhances training stability by implementing constraints on policy modifications. PPO seeks to facilitate stable learning by maintaining updates within a designated trust region, consequently defining the surrogate objective function $J^{CLIP}$, which is optimized to iteratively update the policy $\pi_\theta$:

\begin{equation}
    \label{eq:1}
    J^{CLIP}(\pi_\theta)=\mathbb{E}\left[\min\left(\frac{\pi_\theta(a_t|s_t)}{\pi_{\theta_{old}}(a_t|s_t)}\hat{A}_t,g\left(\epsilon,\hat{A}_t\right)\right)\right],
\end{equation}

\noindent where $\theta_{old}$ is the vector of policy parameters before the update, $\hat{A}_t$ is an estimated advantage value at time $t$, and $g$ is a clipping function defined by:

\begin{equation}
    \label{eq:2}
    g\left(\epsilon,A\right)=\cases{
        \left(1+\epsilon\right)A &for $A \ge 0$,\\
        \left(1-\epsilon\right)A &for $A < 0$,\\
    }
\end{equation}

\noindent where $\epsilon$ is the clipping parameter of PPO and $A$ is an advantage.

\subsection{Intrinsic Rewards}
\label{subsec:intrinsic_rewards}
The exploration-exploitation trade-off has been a challenging problem in RL. The agent is tasked with finding an optimal balance between leveraging its accumulated experience to seek the best policy (exploitation) and probing various episodes to uncover a potentially superior policy (exploration). The application of intrinsic rewards is crucial in encouraging effective exploration, particularly in environments where extrinsic rewards are sparse. This strategy has been identified as beneficial especially for video games that models human curiosity\cite{burda2018exploration, badia2020never, strehl2008analysis, ostrovski2017count, bellemare2016unifying, tang2017exploration, pathak2017curiosity}.

Three different types of intrinsic rewards have been proposed: prediction-based, count-based, and memory-based intrinsic reward functions. Memory-based intrinsic reward methods involve maintaining a record of previously encountered states in memory, promoting exploration for finding novel states by assessing the novelty of the current state in comparison to stored memories. The greater the difference from stored memories, the higher the reward\cite{badia2020never}. However, memory-based methods have to compare the current state with all previous states. Therefore, they are considered resource-intensive approaches that require an extensive memory record.

Count-based intrinsic reward methods are computed by counting how often the agent visits each state. They are used in both tabular settings\cite{strehl2008analysis} and more complex models, including context-tree switching density models\cite{ostrovski2017count}, pseudo-counting\cite{bellemare2016unifying}, and locality-sensitive hashing (LSH) techniques\cite{tang2017exploration}. These methods have the advantage of being simple and easy to implement. However, they may be infeasible for problems of vast sizes, such as those involving chemical space\cite{jo2022leco}.

Prediction-based intrinsic reward methods utilize a neural network model that learns previously visited states to give high rewards for unvisited states. This method often incorporates a forward dynamics model to predict subsequent states from the current state and the current action\cite{burda2018exploration, pathak2017curiosity}. As the predictive model learning progresses, the model has lower prediction errors on states it has already visited compared to states it has not visited. The prediction-based method enables efficient exploration by defining intrinsic rewards based on the prediction error.

\section{Related Works}
\label{sec:related_works}
\subsection{Reinforcement Learning for Molecular Generation}
\label{subsec:reinforcement_learning_for_molecular_generation}
Numerous studies have tackled the RL problem by defining an action space, a state, a policy, and an environment. The action space composed of symbol sets that represent molecular structures, a state space made up of symbol substrings, a policy for predicting the next appropriate symbol (action) to append to the current substring (state) up to a certain length, and an environment that evaluates the completed string, providing rewards based on its properties\cite{choi2023rebadd, kim2022predicting, olivecrona2017molecular, pereira2021diversity}. Policies employ deep neural network models, such as RNNs, to deal with string-based molecular structures. Rewards are allocated proportionally to the molecular structure's target chemical properties or pharmacological efficacy metrics. Policy networks are updated using various policy gradient algorithms, including REINFORCE\cite{choi2023rebadd, kim2022predicting, olivecrona2017molecular} and PPO\cite{thiede2022curiosity}. The generation of a molecular structure with the desired properties yields a high reward, incentivizing the policy network to generate more molecular structures associated with higher rewards.

\subsection{Intrinsic Rewards for Molecular Generation}
\label{subsec:intrinsic_rewards_for_molecular_generation}
RL-based techniques for molecular generation are promising techniques for identifying molecular structures with specific desired chemical properties. However, the immense search space of chemical compounds presents a formidable challenge, often limiting the ability of RL models to generate a diverse array of molecular structures\cite{mokaya2023testing}. Various learning strategies incorporating intrinsic rewards have been developed to allow an agent to explore diverse molecular structures efficiently in the vast chemical space\cite{thiede2022curiosity}.

\subsubsection{Count-based Intrinsic Reward}
\label{subsubsec:count_based_intrinsic_reward}
The count-based intrinsic reward method\cite{thiede2022curiosity} computes intrinsic reward values according to the frequency with which molecular structures are observed during training. This approach uses Morgan's fingerprints\cite{cereto2015molecular} to convert molecular structures into numerical vectors and exploits LSH for efficient tracking of the occurrence frequencies of these structures. The intrinsic reward for encountering a molecular structure $m$ is calculated as follows:

\begin{equation}
    \label{eq:3}
    r_{count}\left(m\right)=\frac{1}{\sqrt{\Gamma\left(LSH\left(MF\left(m\right)\right)\right)}+\epsilon},
\end{equation}

\noindent where $MF$ represents the Morgan fingerprint function, $\Gamma$ represents a function that records the occurrence frequency of hash values derived from MF, and $\epsilon$ is a small positive constant introduced to prevent division by zero. This formulation guarantees that the intrinsic reward for a specific molecule decreases as the molecule is encountered more often, thereby promoting the exploration of new and less frequently observed molecular structures.

\subsubsection{Memory-based Intrinsic Reward}
\label{subsubsec:memory_based_intrinsic_reward}
The memory-based intrinsic reward method utilizes a memory data structure to track molecular structures generated throughout the training process. It penalizes the agent for discovering frequently observed molecular structures\cite{thiede2022curiosity}. This memory is organized as a fixed-size First-in-First-Out (FIFO) buffer. At each step, negative intrinsic rewards are assigned based on the degree of similarity between the current molecular structure and those previously stored in memory. This similarity is quantified using the Tanimoto similarity coefficient, and the intrinsic reward for a molecule is computed as follows:

\begin{equation}
    \label{eq:4}
    r_{memory}\left(m\right)=-\max_{q\in Q}\left\{TS\left(MF\left(m\right),MF\left(q\right)\right)\right\},
\end{equation}

\noindent where $Q$ denotes the fixed-sized memory and $TS$ represents the Tanimoto similarity function. This method penalizes the agent when it generates structures that are highly similar to those already stored in the memory buffer in order to prevent the agent from repeatedly rediscovering similar molecular structures. Therefore, it encourages the agent to explore new and less commonly encountered molecular structures.

\subsubsection{Prediction-based Intrinsic Reward}
\label{subsubsec:prediction_based_intrinsic_reward}
The prediction-based intrinsic reward approach\cite{thiede2022curiosity} uses prediction errors with respect to molecular properties to calculate intrinsic rewards. Traditional strategies in this domain might focus on prediction errors in determining the next state\cite{pathak2017curiosity}. However, such methods sometimes fail to yield insights for uncovering novel molecular structures within a generation context. To address this problem, In \cite{thiede2022curiosity}, the authors proposed the use of intrinsic rewards derived from a model to predict specific molecular properties instead of next state prediction. This strategy is formalized as follows:

\begin{equation}
    \label{eq:5}
    r_{prediction}\left(m\right)={\left\| \hat{\phi}\left(m\right)-\phi\left(m\right) \right\|}_{l},
\end{equation}

\noindent where $\phi$ represents a property predictive oracle, $\hat{\phi}$ denotes a neural network that approximates this oracle's predictions, and ${\left\|\cdot\right\|}_l$ signifies either the L1 or L2 norm. This method aims to directly align the exploration process with the discovery of molecular structures exhibiting desirable properties by emphasizing the discrepancy between predicted and actual molecular characteristics.

\section{Limitations of traditional approaches}
\label{sec:limitations_of_traditional_approaches}
In \cite{thiede2022curiosity}, the authors proposed three new types of intrinsic reward for enhancing goal-directed molecular generation: count-based, memory-based, and prediction-based rewards. Through evaluation with benchmarks, including pLogP, QED, and celecoxib-similarity, their efficacy was examined. However, the proposed methods still face challenges in discovering new and unknown molecules. This is because solutions occasionally becoming trapped in local optima, hindering efficient exploration within vast chemical spaces.

The count-based and memory-based methods, which are considered history-based approaches, use crafted storage mechanisms to store the information of previously visited states to calculate intrinsic rewards. However, they require meticulous attention to the design of state history management. Moreover, these methods often result in an imbalance between exploration and exploitation, and ultimately lead to excessive exploration by prioritizing the exploration of unvisited states.

The prediction-based method, a learning-based approach, employs a deep neural network to extract meaningful features from past states. It leverages the neural network's ability to automatically learn and remember states without requiring manually designed storage solutions and also promotes more efficient exploration. However, this approach cannot effectively encourage exploration as RL training progresses, particularly when the agent needs to escape out of local optima. Specifically, when the neural network exhibits unexpectedly high generalization performance, it may predict low errors even for unvisited states and hinder the agent from exploring the space further.

\subsection{History-based Approach}
\label{subsec:history_based_approach}

The history-based approach leverages a predefined storage mechanism to store information on previously visited states. This method prevents revisits to stored states and similar states by leveraging stored state information and encourages exploration of novel states that are not recorded in the history. This approach is effective in tasks where continuous exploration of novel molecular structures is crucial. However, it is not suitable for tasks that require finding molecules with structural similarities to a specific molecule.

\begin{figure}[t]%
\includegraphics[width=\textwidth]{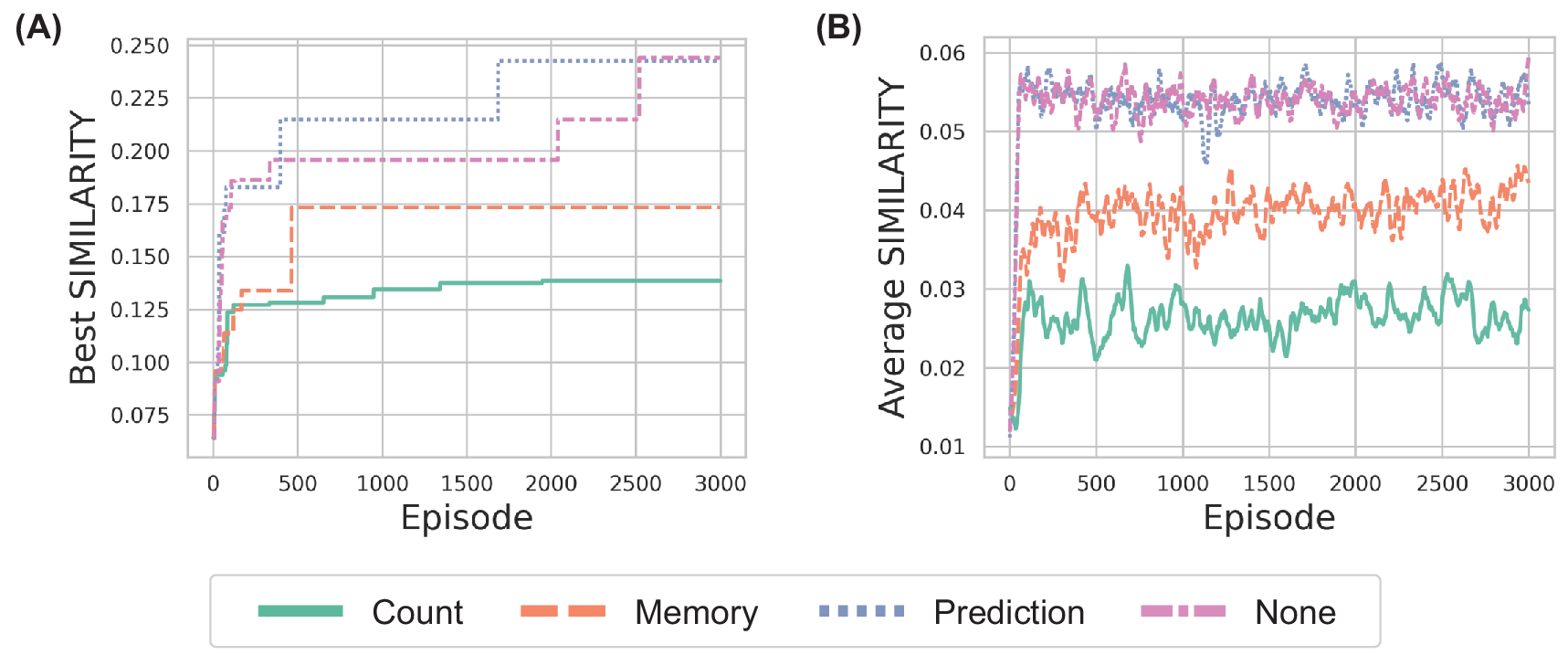}
\caption{
\textbf{\label{fig:1}Comparison of traditional intrinsic reward methods in Celecoxib-like molecular structure generation}. (A) Best similarity scores and (B) average similarity scores over training episodes}
\end{figure}

The main limitation of the existing history-based approaches becomes evident when they are used to discover molecular structures that are similar to specific drugs\cite{thiede2022curiosity}. \Fref{fig:1} shows the structures similar to the drug celecoxib that are discovered across four scenarios: using three types of intrinsic rewards and not using any intrinsic reward.

\Fref{fig:1}A presents the Tanimoto similarity scores of the molecular structures most closely matched to celecoxib that were identified through a previously proposed RL-based framework by \cite{thiede2022curiosity}. \Fref{fig:1}B depicts the average similarity across the discovered molecular structures. The history-based approach, including the count-based and memory-based methods, failed to consistently generate celecoxib-like structures with high Tanimoto similarity. This is primarily because these methods force the agent to find molecular structures that are significantly different from celecoxib, even after identifying similar ones. Therefore, it results in a lower average similarity compared to scenarios even without no intrinsic rewards.

\subsection{Learning-based Approach}
\label{subsec:learning_based_approach}
Learning-based approaches, such as the prediction-based intrinsic reward method, exploit machine learning models to effectively learn previously visited states and calculate intrinsic rewards\cite{stadie2015incentivizing}. Different from history-based approaches, where the state information storage mechanism is predefined, learning-based approaches can dynamically learn and improve this mechanism as the exploration of the agent continues. This feature makes them a versatile tool for goal-directed molecular generation with various target chemical properties. However, the learning-based approach also has limitations when it is solely used. While its high generalizability allows for efficient exploration of an agent, it can also lead to misclassifications where unvisited states are mistakenly identified as visited. This can make it challenging to encourage effective exploration as RL training progresses and can result in the agent getting stuck in local optima.

\begin{figure}[t]%
\includegraphics[width=\textwidth]{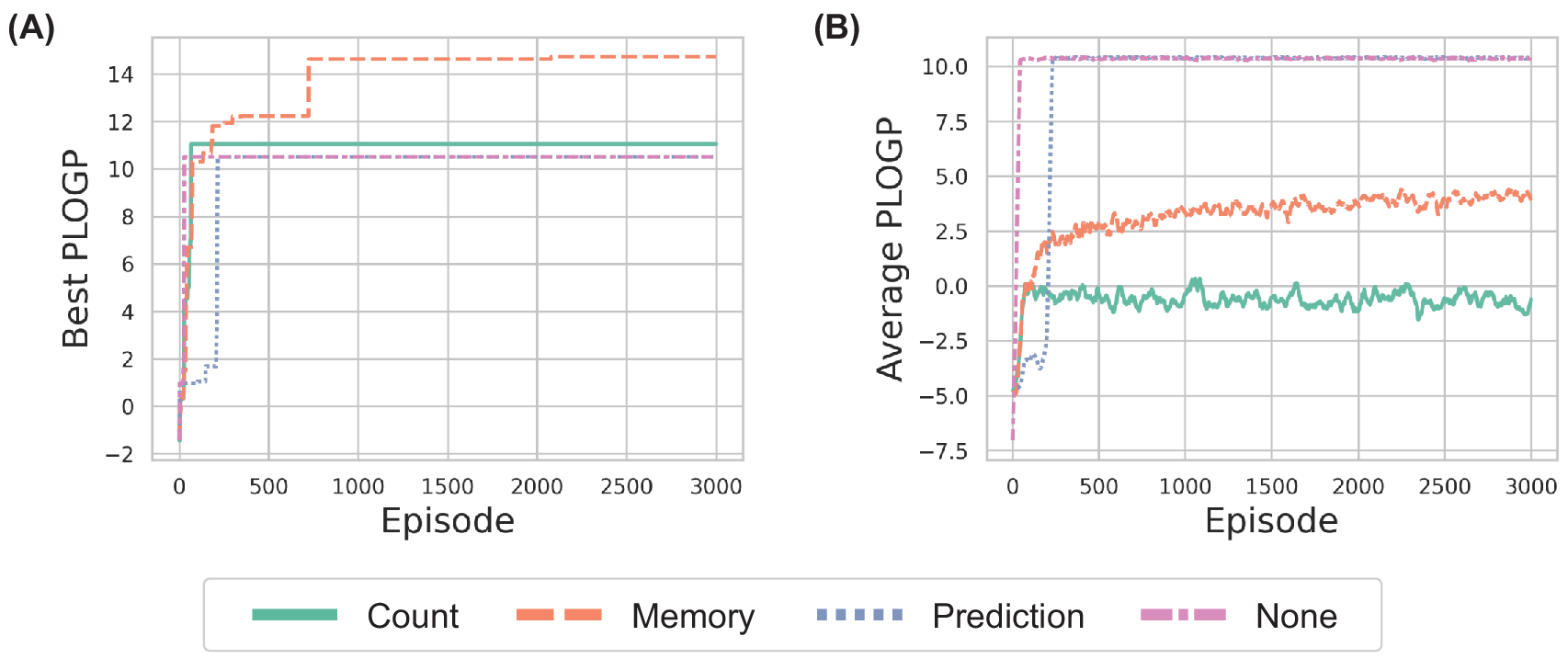}
\caption{\label{fig:2}\textbf{Comparison of traditional intrinsic reward functions in pLogP optimization}. (A) Best pLogP scores and (B) average pLogP scores over training episodes}
\end{figure}

The limitations of the learning-based approach can be observed when it is solely used for pLogP optimization. \Fref{fig:2}A shows the highest pLogP values achieved by various intrinsic reward methods during training.

Two history-based approaches utilizing count-based and memory-based intrinsic rewards, respectively, significantly outperform the case without any intrinsic rewards, achieving consistently higher pLogP scores. However, the learning-based approach utilizing prediction-based intrinsic reward failed to encourage further exploration and discovery of high pLogP molecules. \Fref{fig:2}B shows the average pLogP scores during a training process. The fluctuations of the two history-based approaches indicate that they successfully explored diverse molecule structures and finally discovered high pLogP molecules. On the contrary, the plot for the prediction-based method shows minimal fluctuations, indicating that the agent struggles to explore new molecular structures. This is because it has difficulty in performing further exploration when it tries to escape from local optima.

\begin{figure}[t]%
\includegraphics[width=\textwidth]{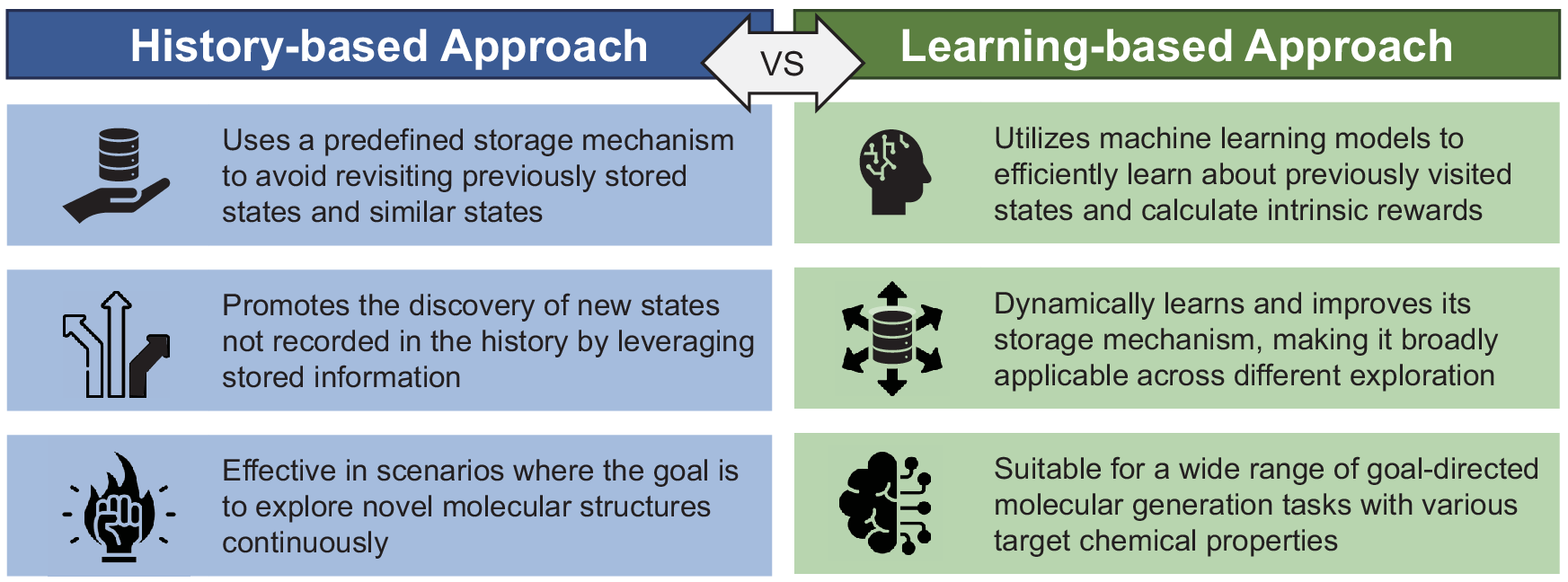}
\caption{\label{fig:3}\textbf{Comparison of history-based and learning-based intrinsic reward approaches}}
\end{figure}

As shown in \fref{fig:3}, relying solely on a history-based or learning-based approach for drug design leads to inefficient exploration in vast chemical spaces. This observation motivates the development of a hybrid approach where the two approaches are synergistically combined to calculate adaptive intrinsic rewards. Our goal is to overcome the current limitations of both history-based and learning-based approaches and finally create a more robust framework that efficiently explores the vast chemical space and optimizes molecular structures.

\section{Methods}
\label{sec:methods}
In this study, we introduce Mol-AIR, a molecular optimization framework with adaptive intrinsic rewards that performs efficient exploration for effective goal-directed molecular generation. Mol-AIR integrates the strengths of both history-based and learning-based intrinsic approaches for efficient exploration. This approach synergizes the history-based intrinsic reward (HIR) with the learning-based intrinsic reward (LIR) to achieve efficient exploration of the molecular structure state space to identify target properties.

HIR facilitates the exploration of the chemical space by counting the number of visits to each state. This module encourages the discovery of new molecular structures by prioritizing less visited states. Concurrently, LIR adjusts the balance between exploration and exploitation through the implementation of an RND method\cite{burda2018exploration}. We introduce the RND for efficient exploration in the vast chemical space, as it is known to be effective for encouraging exploration in sparse reward environments in video games using intrinsic rewards. Moreover by obtaining new information through errors in action prediction, it can increase the value of rewards in sparse reward environments. RND computes intrinsic rewards using the difference of outputs between two neural networks, which enables more efficient exploration of the RL agent. By combining HIR and LIR, Mol-AIR provides a powerful framework for navigating the vast and complex landscape of molecular structures by providing intrinsic rewards. This enables the efficient identification of molecules with desired chemical properties.

\begin{figure}[t]%
\includegraphics[width=\textwidth]{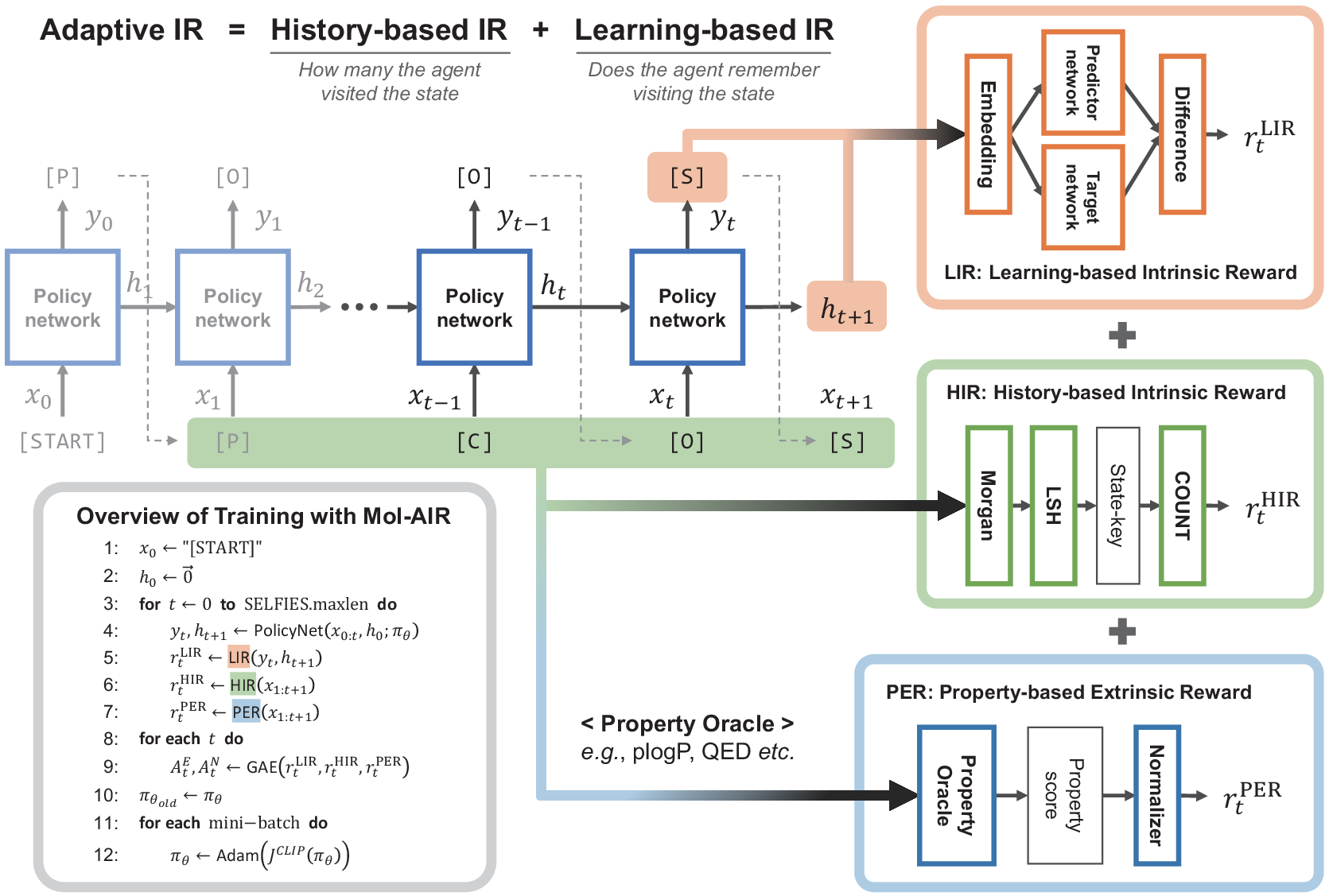}
\caption{\label{fig:4}\textbf{Overview of Mol-AIR}}
\end{figure}

\Fref{fig:4} illustrates the process of calculating two intrinsic rewards and one extrinsic reward in the training of RL-based models with Mol-AIR. The descriptions of the RL environment are provided in the Supplementary Information. The molecular structure generator (policy) completes the molecular structure by predicting and appending SELFIES characters one by one and calculates the count-based intrinsic reward, RND-based intrinsic reward, and extrinsic reward based on target properties each time a SELFIES character is added. The three rewards are then used to update the policy network through the PPO algorithm, and upon completion of the training, the policy network gains the ability to output molecular structures optimized for the target properties. The specific details are given in the subsections that follow.

\subsection{HIR with Exponential Decay over Visit Counts}
\label{subsec:hir_with_exponential_penalty_on_counts}
In this study, we propose a new count-based intrinsic reward method to efficiently record information about previously visited states. The proposed method tracks the number of times each molecule is visited and also encourages exploration of molecule structures that have not been seen before or have been encountered rarely by assigning high intrinsic rewards to those states. We utilize Morgan fingerprints \cite{cereto2015molecular} to capture molecular information and the LSH algorithm to facilitate efficient record-keeping of previously visited structures. Specifically, at time step $t$, the value of HIR, denoted as $r_{t}^{HIR}$, is calculated as follows:

\begin{equation}
    r_{t}^{HIR}:=\exp\left[-\min\left\{\tau,\Gamma\left(LSH\left(MF\left(x_{1:t+1}\right)\right)\right)\right\}\right],
    \label{eq:6}
\end{equation}

\noindent where $x_{1:t+1}$ represents a molecular structure consisting of $t+1$ of SELFIES characters and $\tau$ is an upper bound of the frequency of occurrence. In this study, $\tau$ was set to 10.

\subsection{LIR with Random Network Distillation}
\label{subsec:lir_with_random_network_distillation}
In this study, we employ an RND-based approach to leverage the advantages of adaptability and flexibility inherent in learning-based approaches. This method involves learning the information of visited states using two neural networks (Supplementary Figure S1) and calculating the intrinsic reward based on the difference in predictions between these two networks. The RND-based method randomly initializes two neural networks with similar architectures and trains only one network to make its outputs identical to those of the other network when given state information as an input. The network being trained is referred to as the predictor network, while the network that retains its initial state is called the target network. As the training progresses, the difference in the results of the two models for the visited state information decreases, allowing the definition of an intrinsic reward that encourages exploration towards unvisited states using the error in results. Specifically, the intrinsic reward based on RND at time $t$, $r_{t}^{LIR}$, is calculated as follows:

\begin{equation}
    r_{t}^{LIR}:=\frac{1}{2\sigma^R}{\left\|\mathcal{F}\left(y_t,h_{t+1}\right)-\mathcal{F}_\phi\left(y_t,h_{t+1}\right)\right\|}^2,
    \label{eq:7}
\end{equation}

\noindent where $\mathcal{F}$ represents the target network, $\mathcal{F}_\phi$ represents the predictor network, and ${\sigma}^R$ is the running standard deviation of the RND-based intrinsic returns, which is used to reduce the scale difference of intrinsic rewards over time. Training of the predictor network is carried out alongside policy training, and the mean squared error between the outputs of the two neural networks is used as a loss function.

\subsection{PER with Objective Property Oracle}
\label{subsec:per_with_objective_property_oracle}
In molecular structure generation tasks, the goal of reinforcement learning is to find molecular structures with superior targeted chemical properties. Once the policy selects an action to append a SELFIES character to the SELFIES string constructed so far, the environment evaluates the chemical properties of the current molecular structure and produces an extrinsic reward based on the evaluation score to update the policy, thereby achieving the goal. Specifically, given an oracle $\Lambda^{(p)}$ that can evaluate the target chemical property $p$, the value of PER at time step $t$, $r_{t}^{PER}$ is calculated as follows:

\begin{equation}
     r_{t}^{PER}:=\Lambda^{(p)}\left(x_{1:t+1}\right) - \Lambda^{(p)}\left(x_{1:t}\right),
     \label{eq:8}
\end{equation}

\noindent As SELFIES substrings can always be converted to structurally valid SMILES strings\cite{krenn2022selfies}, in this study, we calculated the extrinsic reward at every time step and reflected it in the policy update.

\subsection{Two Types of Advantage Estimation}
\label{subsec:two_types_of_advantage_estimation}
To effectively learn two intrinsic rewards and one extrinsic reward, this study employs an actor-critic structure (Supplementary Figure S2). Following the approach proposed by \cite{burda2018exploration}, we use two critic networks to estimate the values of episodic and non-episodic advantages at each time step. The episodic advantage encourages exploration and exploitation progress within episodes, while the non-episodic advantage encourages exploration throughout the entire learning process.

The episodic critic network learns both the property-based extrinsic reward and the count-based intrinsic reward and calculates the episodic state-value $V_{\omega}^{E}(x_t,h_t)$ at time step $t$. The episodic advantage $A_{t}^{E}$ at time step $t$ is then calculated using the generalized advantage estimation (GAE)\cite{schulman2015high} as follows:

\begin{equation}
    A_{t}^{E}=\sum_{l=t}^{L}{\left(\gamma\lambda\right)^{l-t}\left(r_{l}^{PER}+\alpha{\beta}r_{l}^{HIR}+\gamma{V_{l+1}^E}-{V_{l}^{E}}\right)},
    \label{eq:9}
\end{equation}

\noindent where $L$ is the maximum length of SELFIES, $\gamma$ is a discount factor, $\lambda$ is a bias-variance trade-off parameter, $\alpha$ is a weight of intrinsic rewards, and $\beta$ is a parameter for balancing between HIR and LIR.

The non-episodic critic network learns the RND-based intrinsic reward and calculates the non-episodic state-value $V_{\psi}^{N}(x_t,h_t)$ at time step $t$. The non-episodic advantage $A_{t}^{N}$ at time step $t$ is calculated using the GAE as follows:

\begin{equation}
    A_{t}^{N}=\sum_{l=t}^{\infty}{\left(\gamma\lambda\right)^{l-t}\left(\alpha{r_{l}^{LIR}}+\gamma{V_{l+1}^{N}}-{V_{l}^{N}}\right)},
    \label{eq:10}
\end{equation}

\subsection{Policy Gradient with AIR and PER}
\label{subsec:policy_gradient_with_air_and_per}
The actor network $\pi_\theta$, corresponding to the policy, is trained using the sum of two advantage values ${A}_{t}^{E}+{A}_{t}^{N}$ and the PPO-Clip algorithm. Specifically, for a given current SELFIES character $x_t$, the previous hidden state $h_t$, and the next SELFIES character $y_t$, an objective function $J^{CLIP}\left(\pi_\theta\right)$ is calculated, and the actor network is updated using the policy gradient algorithm. Algorithm \ref{alg:1} illustrates the training algorithm of the proposed model and some model parameters are described in Supplementary Table S1.

\begin{algorithm}
\caption{\label{alg:1}Training algorithm with Mol-AIR}
\begin{algorithmic}
\Require A family of sets of initial weights for actor, two critic and RND networks $\{\theta, \omega, \psi, \phi\}$, the number of episodes $N$, the number of environments $M$, and the maximum length of SELFIES $L$
\Ensure A trained policy $\pi_{\theta}$
\For{$n = 1, \cdots, N$}
    \State{Initialize a replay buffer $\mathcal{B}$}
    \For{$m = 1,\cdots, M$}
    
        \State{$x_0\gets$ ``[START]''} \Comment{Set an initial token}
        \State{$h_0\gets \vec{0}$} \Comment{Set an initial hidden state with a zero vector}
        
        \For{$t = 0, \cdots, L-1$}
            \State{$p_t, h_{t+1} \leftarrow \pi_\theta(x_t, h_t)$}
            \State{$y_t \sim p_t(y)$} \Comment{Sample the next SELFIES character}
            \State{calculate a history-based intrinsic reward $r_t^{HIR}$} \Comment{\Eref{eq:6}}
            \State{calculate a learning-based intrinsic reward $r_t^{LIR}$} \Comment{\Eref{eq:7}}
            \State{calculate a property-based extrinsic reward $r_t^{PER}$} \Comment{\Eref{eq:8}}
            \If{$y_t =$ ``[END]''}
                \State{break}
            \EndIf
        \EndFor    
    
        \For{\textbf{each} time step $t$}
            \State{calculate episodic advantage ${A}_{t}^{E}$ using ${V}_{\omega}^{E}$} \Comment{\Eref{eq:9}}
            \State{calculate non-episodic advantage ${A}_{t}^{N}$ using ${V}_{\psi}^{N}$} \Comment{\Eref{eq:10}}
        \EndFor

        \State{store all of experiences into the buffer $\mathcal{B}$}
    
    \EndFor
    \State{${\pi}_{{\theta}_{old}} \gets {\pi}_{\theta}$} \Comment{Keep the old policy for \Eref{eq:1}}
    \For{\textbf{each} mini-batch sampled from $\mathcal{B}$}
        \State{optimize $\theta$ wrt PPO loss with Entropy regularization using Adam}
        \State{optimize $\omega$ and $\psi$ wrt critic loss using Adam}
        \State{optimize $\phi$ wrt RND loss using Adam}
    \EndFor
\EndFor
\end{algorithmic}
\end{algorithm}

\section{Results}
\label{sec:results}
\subsection{Implementation Details}
\label{subsec:implementation_details}
The Mol-AIR methodology was implemented in Python 3.7, taking advantage of a suite of open-source tools to facilitate the development and evaluation of the molecular generative model. The key libraries and frameworks used in this implementation included PyTorch 1.11.0 for deep learning models, CUDA 11.3 to harness GPU acceleration, RDKit 2022.9.5 and SELFIES 0.2.4 for cheminformatics support and robust molecular string representation, respectively, and PyTDC 0.4.0 for assessing the chemical properties of generated molecules. The computational experiments were conducted on an Ubuntu 20.04.6 LTS system equipped with 251 GiB of memory and a NVIDIA RTX A6000 GPUs.

\subsection{Benchmark Tasks}
\label{subsec:benchmark_tasks}
To demonstrate the superiority of the proposed intrinsic reward method, we set six target properties as benchmarks and searched for the optimal molecular structure for each property. Among the six target properties used in this benchmark test, three (pLogP, QED, similarity to the molecule celecoxib) were adopted from previous studies\cite{brown2019guacamol, thiede2022curiosity}, and the remaining three (GSK3B, JNK3, GSK3B+JNK3) are widely used molecular properties in goal-directed molecular generation studies\cite{li2018multi, fromer2023computer}. The descriptions of their properties are as follows:

\begin{itemize}
    \item \textbf{pLogP}: The penalized logarithm of the octanol-water partition coefficient simultaneously assesses both the hydrophobicity of a molecule and its chemical feasibility\cite{gomez2018automatic}. The range of the pLogP score lies between $\left(-\infty,\infty\right)$, and the goal of pLogP optimization in this study is to find a molecular structure with a high pLogP score. Based on the previous study\cite{thiede2022curiosity}, we normalized pLogP scores by multiplying them by 0.1.
    
    \item \textbf{QED}: The QED measures the likelihood that a molecule can be used as a drug, considering eight physicochemical properties\cite{bickerton2012quantifying}. The range of the QED score is on the half-open interval $\left[0,1\right)$, and the goal of the QED optimization task is to find a molecule with a high QED score.
    
    \item \textbf{Similarity}: This task was designed to evaluate the rediscovery performance of de novo molecular design methods in the Guacamol benchmark\cite{brown2019guacamol}. The goal of this similarity task is to find a molecule that is similar to celecoxib in terms of Tanimoto similarity of Morgan fingerprints. The range of Tanimoto similarity is in the closed interval $\left[0,1\right]$.
    
    \item \textbf{GSK3B}: This property measures the inhibitory ability of a molecule against glycogen synthase kinase-3 beta (GSK3B), a potential therapeutic target for Alzheimer’s disease (AD) due to its association with AD pathophysiology\cite{lauretti2020glycogen}. The GSK3B score is on the closed interval $\left[0,1\right]$, and the goal of this task is to find a molecule with a high GSK3B score.
    
    \item \textbf{JNK3}: This property measures the inhibitory ability of a molecule against c-Jun N-terminal kinase 3 (JNK3), considered a drug target for AD treatment because its overexpression has been found to induce cognitive deficiency\cite{solas2023jnk}. The JNK3 score is on the closed interval $\left[0,1\right]$, and the goal of this task is to find a molecule with a high JNK3 score.
    
    \item \textbf{GSK3B+JNK3}: A previous study suggested that a molecule capable of inhibiting both GSK3B and JNK3 simultaneously could be a potential drug candidate for AD treatment\cite{li2018multi}. To find molecular candidates for GSK3B and JNK3 dual inhibitors, we set the arithmetic mean of GSK3B and JNK3 scores as an objective score and aim to find a molecule with a high mean score.
\end{itemize}

For the calculation of pLogP, QED, and similarity scores, we used scripts provided by existing research\cite{thiede2022curiosity}, and for GSK3B and JNK3 scores, we utilized the oracle provided by the Therapeutics Data Commons (TDC) library\cite{huang2022artificial}.

\subsection{Molecular Discovery with the Best Property Scores}
\label{subsec:molecular_discovery_with_the_best_property_scores}
We compared our framework with the baseline methods that utilize three intrinsic rewards\cite{thiede2022curiosity}. In the performance measurement experiments for the baseline techniques, the model parameter values for pLogP, QED, and similarity were set according to the values reported in \cite{thiede2022curiosity}, and for GSK3B and JNK3, optimal model parameters were determined using a grid search approach (Supplementary Table S2). All experimental results were evaluated based on 3,000 iterations of training.

\begin{table}
    \caption{\label{tab:1}Results of Molecular Discovery with the Best Property Scores}
    \begin{indented}
    \item[]\begin{tabular}{lcccccc}
        \br
                    &  \multicolumn{6}{c}{Best Property Score} \\
         Intrinsic  &  \crule{6} \\
         Type       & pLogP    & QED      & Similarity    & GSK3B       & JNK3        & GSK3B+JNK3    \\
         \mr
         None       & 10.523   & 0.898    & 0.244         & 0.490       & 0.240       & 0.285         \\
         Count      & 11.063   & 0.869    & 0.139         & 0.407       & 0.260       & 0.250         \\
         Memory     & 14.740   & 0.931    & 0.173         & 0.440       & 0.240       & 0.285         \\
         Prediction & 10.523   & 0.919    & 0.243         & 0.480       & 0.250       & 0.313         \\
         \mr
         Mol-AIR    & \textbf{15.572}   & \textbf{0.948}    & \textbf{0.330}         & \textbf{0.730}       & \textbf{0.480}       & \textbf{0.420}         \\
         \br
    \end{tabular}
    \end{indented}
\end{table}

\Tref{tab:1} shows that the proposed method, Mol-AIR, is superior to the baseline approaches, outperforming them in finding the best scoring molecules for all six tasks. In particular, in the search for molecular structures similar to celecoxib, where existing methods failed to do so, the proposed framework was able to create molecular structures with a similarity that exceeded 30\%. Furthermore, the existing baseline methods failed to reach 0.948, the theoretical optimum of QED, but the proposed Mol-AIR succeeded in reaching the theoretical optimum through efficient exploration.

\begin{figure}[t]%
    \includegraphics[width=\textwidth]{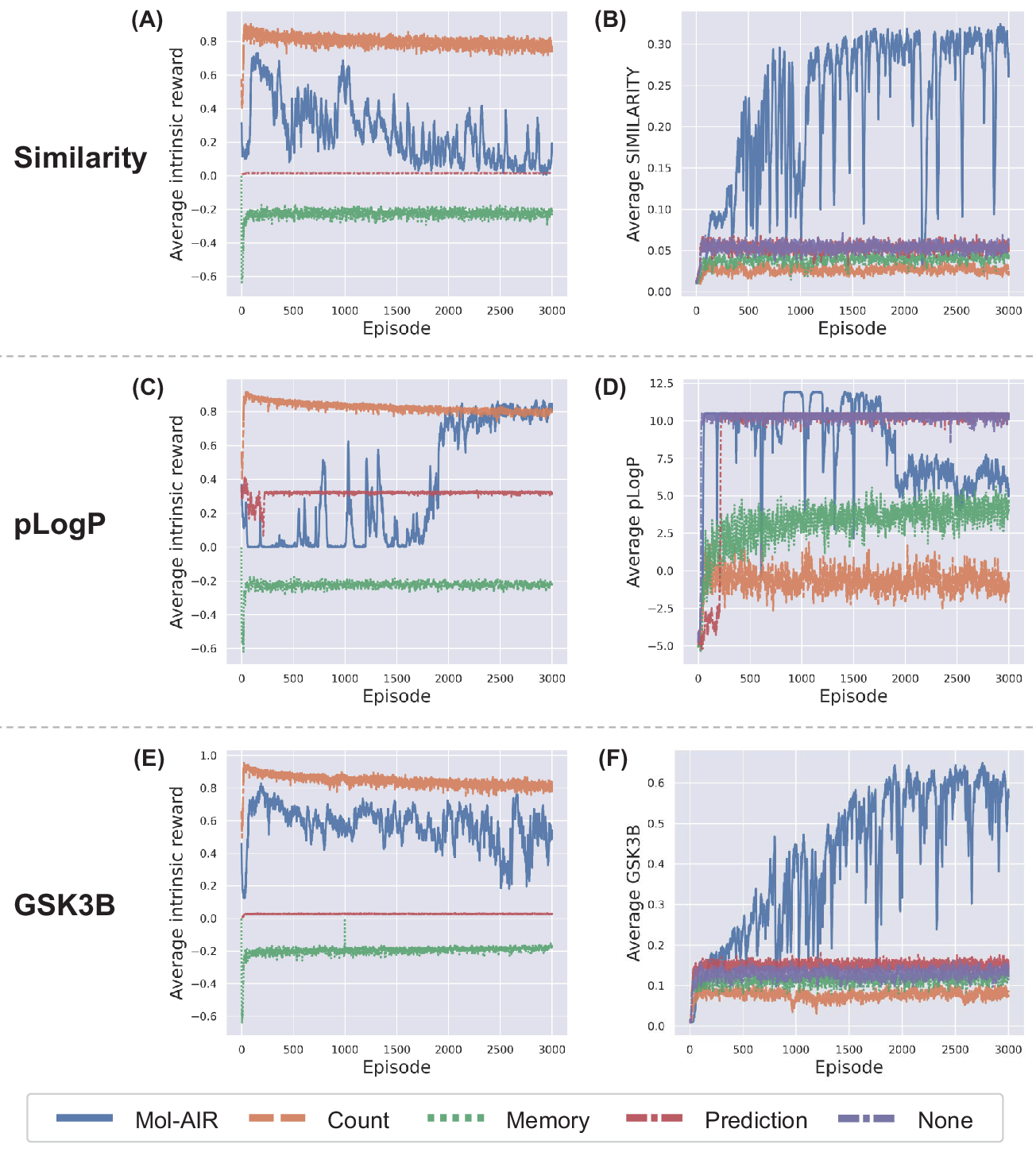}
    \caption{\label{fig:5}\textbf{The Average intrinsic reward and Average objective property score of the molecules generated per batch over the training episodes.} Each row has two line-plots showing the changes of average intrinsic rewards (left) and average property scores (right) over episodes.}
\end{figure}

To analyze the characteristics of the proposed method, we generated 64 molecular structures at every episode, investigating their average intrinsic rewards and average target property values (\Fref{fig:5}).

Mol-AIR, our proposed method, was able to find molecule structures similar to the target molecule because we designed an appropriate adaptive intrinsic reward function that promotes consistent and efficient exploration. Existing methods lack this function. This reward function effectively induces exploration by combining the strengths of history-based and learning-based approaches. It gradually decreases the intrinsic reward after finding structures similar to celecoxib, and thus exploration and exploitation are effectively balanced, which lead to the successful discovery of desirable molecular structures based on extrinsic rewards (\Fref{fig:5}A—\Fref{fig:5}B).

In the case of pLogP, we confirmed that the proposed method provided small intrinsic rewards when exploitation was needed (episode $<$ 1500) and large intrinsic rewards when exploration was needed (episode $>$ 1500) (\Fref{fig:5}C—\Fref{fig:5}D). For GSK3B, Mol-AIR showed large variance in intrinsic rewards, which implies it adaptively balances between exploration and exploitation to find optimal molecular structures (\Fref{fig:5}E). This adaptive balancing led to the continuous growth of the target property scores and thus Mol-AIR achieved a significantly high score of 0.730 compared to the baseline methods (\Fref{fig:5}F). Results for the other target properties are provided in the Supplementary Figure S3. 

These results demonstrate that the proposed method can provide appropriate intrinsic rewards considering the target property and the progress of training, thereby enabling effective training of the molecular structure generation model.

\begin{figure}[t]%
\includegraphics[width=\textwidth]{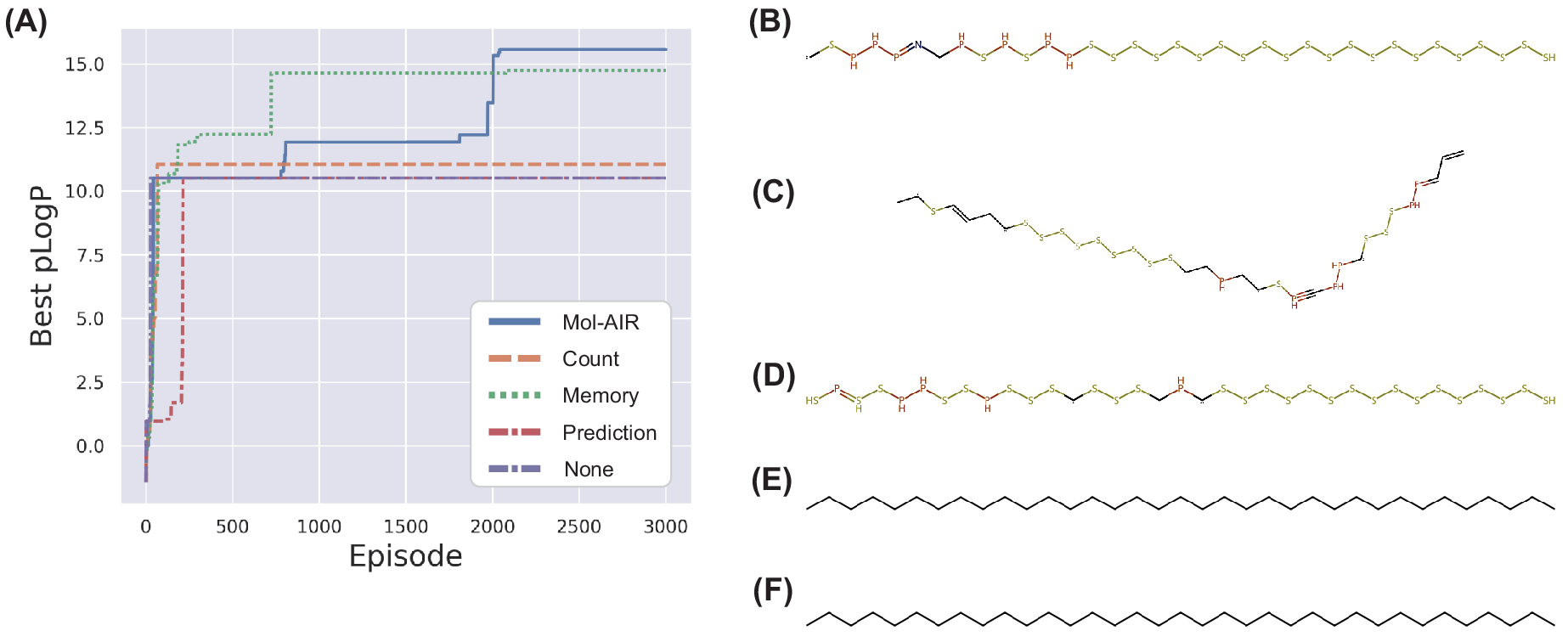}
\caption{\label{fig:6}\textbf{Results of the pLogP benchmark task.} (A) The pLogP value of the best generated molecules in the training run over episodes. The molecules with the highest pLogP value discovered by using (B) Mol-AIR, (C) Count-, (D) Memory-, (E) Prediction-based intrinsic rewards, and (F) no intrinsic reward.}
\end{figure}

Moreover, the proposed method successfully discovered a sulfur-phosphorus-nitrogen chain with a higher pLogP value than the sulfur chain found in previous research (\Fref{fig:6}).

As shown in \fref{fig:6}A, the proposed method initially discovered carbon chains similar to those discovered with prediction-based intrinsic rewards and no intrinsic rewards, found a sulfur chain after training for 750 episodes, and discovered a sulfur-phosphorus-nitrogen chain that was superior to the sulfur-phosphorus chain found with a memory-based intrinsic reward after 2,000 episodes of training. \Fref{fig:5}C also shows that the proposed method significantly increased its exploration capability to approximately 2,000 episodes, achieving higher scores, demonstrating that the proposed method can provide superior exploratory power over existing methods. The results for the other properties are shown in Supplementary Figures S4—S9.

\subsection{Ablation Study}
\label{subsec:ablation_study}
We performed ablation experiments in six benchmark tests to demonstrate the benefits of combining HIR and LIR. \Tref{tab:2} presents the results of the generation of molecular structures with the highest scores for each benchmark test. It demonstrates that the Mol-AIR method, which uses both HIR and LIR, outperforms others in all benchmarks. The use of either HIR or LIR alone decreased performance in all cases, with HIR particularly proving to be more effective than LIR.

\begin{table}
    \caption{\label{tab:2}Results of Ablation Study with the Best Property Scores}
    \begin{indented}
    \item[]\begin{tabular}{lcccccc}
        \br
                    &  \multicolumn{6}{c}{Best Property Score} \\
         Intrinsic  &  \crule{6} \\
         Type       & pLogP    & QED      & Similarity    & GSK3B       & JNK3        & GSK3B+JNK3    \\
         \mr
         None       & 10.523   & 0.898    & 0.244         & 0.490       & 0.240       & 0.285         \\
         HIR only   & 11.930   & 0.916    & 0.296         & 0.680       & 0.270       & 0.395         \\
         LIR only   & 10.523   & 0.922    & 0.298         & 0.650       & 0.250       & 0.360         \\
         \mr
         Mol-AIR    & \textbf{15.572}   & \textbf{0.948}    & \textbf{0.330}         & \textbf{0.730}       & \textbf{0.480}       & \textbf{0.420}         \\
         \br
    \end{tabular}
    \end{indented}
\end{table}

\begin{figure}[t]%
\includegraphics[width=\textwidth]{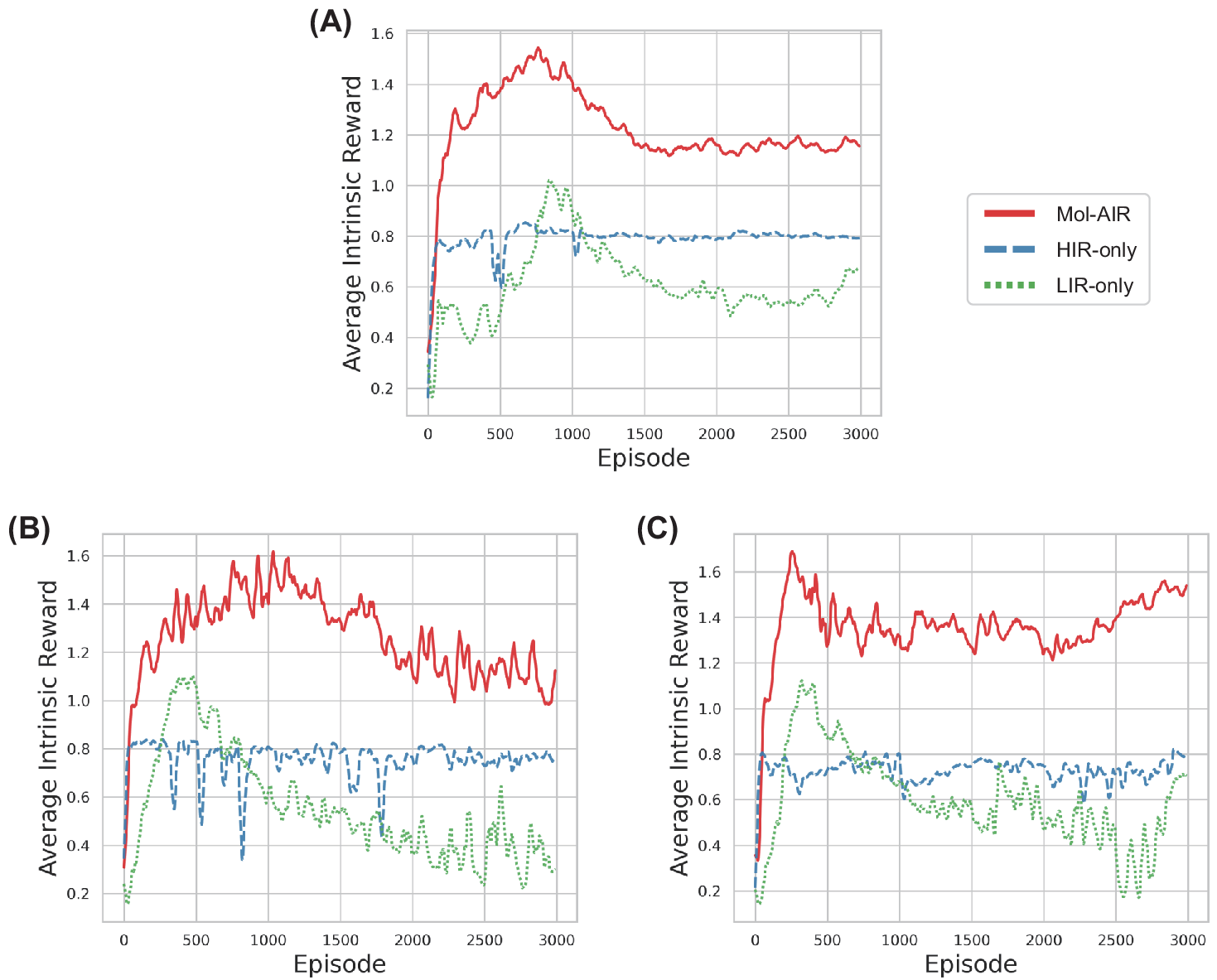}
\caption{\label{fig:7}\textbf{Comparison of intrinsic reward patterns during training among the ablation cases.} (A) QED, (B) GSK3B, and (C) GSK3B+JNK3}
\end{figure}

To investigate why AIR outperforms the singular application of HIR and LIR, we analyzed intrinsic reward patterns across three benchmark tests: QED, GSK3B, and GSK3B+JNK3. The analysis revealed distinct patterns for AIR, LIR, and HIR (\Fref{fig:7}). LIR failed to enhance exploration owing to low intrinsic rewards at the start of training but showed an increase in reward magnitude as the neural network progressed and remembered visited states. However, as the prediction accuracy of the neural network model improved for unvisited states through learning and generalization, the trend of increasing intrinsic rewards reversed, leading to a decrease. This pattern indicated that LIR, except at the beginning, was ineffective in inducing long-time exploration.

On the contrary, HIR provided a consistent level of intrinsic reward from the start, as it does not require a learning process. However, its constant encouragement for exploration made exploitation difficult as training progressed. The AIR pattern proposed in this study adequately combines the HIR and LIR patterns. Specifically, similar to LIR, it produces high intrinsic rewards at the beginning of RL to induce strong exploration and then provides steady intrinsic rewards like HIR, assisting RL by allowing attempts at new structures towards the end. Results for other target properties also show that HIR and AIR work together to facilitate appropriate exploration (Supplementary Figure S10). These results confirm that combining HIR and LIR, as in AIR, is more effective in encouraging exploration than using either alone.

\begin{table}
    \caption{\label{tab:3}Results of Hyperparameter Analysis with Ranking Scores}
    \begin{indented}
    \item[]\begin{tabular}{lcccc}
        \br
         $\beta$          & 1.0         & 0.1     & 0.01          & 0.001       \\
         \mr
         plogP            & 3           & 2       & \textbf{1}    & 4           \\
         QED              & 4           & 2       & \textbf{1}    & 3           \\
         Similarity       & 4           & 3       & \textbf{1}    & 2           \\
         GSK3B            & 4           & 3       & 2             & \textbf{1}  \\
         JNK3             & \textbf{1}  & 3       & 2             & 4           \\
         GSK3B+JNK3       & 4           & 3       & 2             & \textbf{1}  \\
         \mr
         Average Ranking ($\downarrow$)  & 3.3         & 2.7     & \textbf{1.5}  & 2.5         \\
         \br
    \end{tabular}
    \end{indented}
\end{table}

\subsection{Hyperparameter Analysis}
\label{subsec:hyperparameter_analysis}
For Mol-AIR to achieve optimal exploration, it is crucial to adjust the weight of intrinsic rewards $\alpha$ and the HIR-LIR balancing parameter $\beta$ in the equations \eref{eq:9}—\eref{eq:10}. As the pattern of extrinsic rewards varies depending on the target property, $\alpha$ requires a heuristic setting. However, $\beta$, which controls the exploratory dominance between HIR and LIR, was investigated for the optimal ratio between various target properties with values of $\beta$ in the set $\left\{1, 0.1, 0.01, 0.001\right\}$. Supplementary Table S3 presents the results of comparative experiments on $\beta$ for six target properties, and \tref{tab:3} shows the ranking results. It was observed that a value of $\beta$ of 0.01 statistically outperforms across multiple target properties. As the scale of HIR is approximately 100 times larger than that of LIR, setting $\beta$ to 0.01 seems to best balance HIR and LIR, preventing the dominance of excessively large intrinsic rewards and leading to a harmonious exploration between HIR and LIR.

\section{Conclusion}
\label{sec:conclusion}
In this study, we propose a new RL-based framework based on a novel intrinsic reward that performs efficient exploration for goal-directed molecular generation. The proposed approach is a hybrid approach for effective exploration, utilizing an adaptive intrinsic reward function that combines the strengths of history-based and learning-based approaches. Our results reveal that this approach is effective for efficient exploration in the chemical space, as it successfully discovers molecules that are better than compounds discovered by existing intrinsic reward methods. An ablation study revealed that the proposed method's two components, LIR and HIR, synergistically contribute to its success. LIR facilitates strong early-phase exploration, while HIR ensures sustained exploration later on, ultimately guiding the RL agent towards optimal molecular structures. Mol-AIR successfully performs efficient exploration, balancing exploration and exploitation for goal-directed molecular generation. As a result, we discovered new desired molecular structures for various target chemical properties.

Although the proposed technique showed superior performance in the benchmark test for discovering molecular structures similar to celecoxib compared to existing intrinsic reward methods, it still exhibited low similarity. Although AIR adaptively calculates intrinsic rewards and regulates exploratory power considering visited states, tasks sensitive to the exploration of new structures, such as discovering similar molecular structures, require more refined control of exploration. As the proposed AIR and existing methods calculate intrinsic rewards independently of target property information, fine-tuning exploration is challenging, indicating a need for the improvement of AIR so that it can consider extrinsic rewards.

The technology for generating molecular structures similar to existing drugs is a crucial technique that can be utilized in lead optimization for drug development. In future work, we will focus on developing effective intrinsic rewards and RL techniques for training models capable of generating structurally similar molecular structures.

\section*{CRediT authorship contribution statement}
\textbf{Jinyeong Park:} Methodology, Software, Writing - Original Draft. \textbf{Jaegyoon Ahn:} Conceptualization, Methodology. \textbf{Jonghwan Choi:} Validation, Investigation, Writing - Review \& Editing. \textbf{Jibum Kim:} Supervision, Project administration, Funding acquisition.

\section*{Data Availability}
Our data and code are available at https://github.com/DevSlem/Mol-AIR.

\section*{Declaration of competing interest}
The authors declare that they have no known competing financial interests or personal relationships that could have appeared to influence the work reported in this paper.

\ack
This work was supported in part by the National Research Foundation of Korea (NRF) Grant funded by the Korean Government (MSIT) under Grant NRF-2022R1A4A5034121. This work was also supported in part by the MSIT (Ministry of Science and ICT), Korea, under the ICAN (ICT Challenge and Advanced Network of HRD) support program (IITP-2024-RS-2023-00260175) supervised by the IITP (Institute for Information \& Communications Technology Planning \& Evaluation)

\section*{References}
\bibliographystyle{iopart-num}
\providecommand{\newblock}{}

\end{document}